\pdfoutput=1

\documentclass[11pt]{article}

\usepackage{acl}
\usepackage[table]{xcolor}
\usepackage{times}
\usepackage{latexsym}

\usepackage{enumitem} 
\usepackage{subcaption}

\usepackage[T1]{fontenc}

\usepackage[utf8]{inputenc}

\usepackage{microtype}

\usepackage{inconsolata}

\usepackage{graphicx}
\usepackage{amsmath}
\title{SemShareKV: Efficient KVCache Sharing for Semantically Similar Prompts via Token-Level LSH Matching}

\author{Xinye Zhao \\
  University of Notre Dame\\
  \texttt{xzhao24@nd.edu} \\\And
  Spyridon Mastorakis \\
  University of Notre Dame\\
  \texttt{smastor2@nd.edu} \\}

\begin{document}
\maketitle
\begin{abstract}
As large language models (LLMs) continue to scale, the memory footprint of Key-Value (KV) caches during inference has become a significant bottleneck. Existing approaches primarily focus on compressing KV caches within a single prompt or reusing shared prefixes or frequently occurred text segments across prompts. However, such strategies are limited in scenarios where prompts are semantically similar but lexically different, which frequently occurs in tasks such as multi-document summarization and conversational agents. We propose \textit{\textbf{SemShareKV}}, a KV cache sharing and compression framework that accelerates LLM inference by reusing KV cache in semantically similar prompts. Instead of relying on exact token matches, SemShareKV applies fuzzy token matching using Locality-Sensitive Hashing (LSH) on token embeddings and incorporates Rotary Position Embedding (RoPE) to better preserve positional information. By selectively reusing relevant KV pairs from a reference prompt's cache, SemShareKV reduces redundant computation while maintaining output quality. Experiments on diverse summarization datasets show up to 6.25$\times$ speedup and 42\% lower GPU memory usage with 5k tokens input, with negligible quality degradation. These results highlight the potential of semantic-aware cache sharing for efficient LLM inference. 


\end{abstract}

\section{Introduction}

Large Language Models (LLMs) have exhibited a strong capability to understand and process human languages, and have been shown to perform comparably to humans in several fields, such as math inference, text memorization, information extraction, story telling~\citep{naveed2023comprehensive}. Recently released LLMs have significantly advanced in processing and comprehending extremely long prompts. However, this  introduces a notable challenge: increased computational demand due to the quadratic time complexity of their Decoder-Only Transformer architecture when handling lengthy text sequences. The issue is further compounded during inference, as the auto-regressive decoding process repeats the computation for each newly generated token~\citep{luohe2024keep}. 

Existing KV cache optimization approaches primarily focus on single-prompt compression through various techniques: \citep{yang2024pyramidinfer} leverage decaying KV importance across layers for selective extraction (though with limited small-batch gains), \citep{gim2024prompt} employ restrictive markup schemas for text chunk reuse, and \citep{yao2024cacheblend} propose deviation-based recomputation that requires impractical per-chunk precomputation for long inputs. Crucially, these methods operate within the constrained paradigm of single-prompt optimization, failing to exploit the substantial efficiency potential of cross-prompt cache reuse, a significant oversight given the prevalence of semantically similar queries in real-world applications where shared computational savings could be substantial.

Motivated by this challenge, we aim to address the following research question: \textbf{\textit{Can we reuse the precomputed KV cache for prompts that are semantically similar?} }


To answer this question, we proposed \textbf{\textit{SemShareKV}}, a KV cache framework that can reuse the cache from one prompt for another that is semantically similar to each other via fuzzy token match. It speeds up prefill phase and compress KV cache in memory. We show that our method can reduce the pre-fill phase time by 6.25$\times$ and save 42\% GPU memory space. We make the following contributions.
\begin{itemize}
    \item We introduce SemShareKV, which explores KV cache sharing across semantically similar prompts based on fuzzy token match.
    \item We evaluate SemShareKV across multiple datasets, demonstrating its effectiveness in accelerating the prefill phase while simultaneously reducing KV cache size.
    \item We explored the role of position encoding in KV cache by injecting it into vector embeddings.
\end{itemize}

\begin{figure*}[t]
  \centering
  \includegraphics[width=\linewidth]{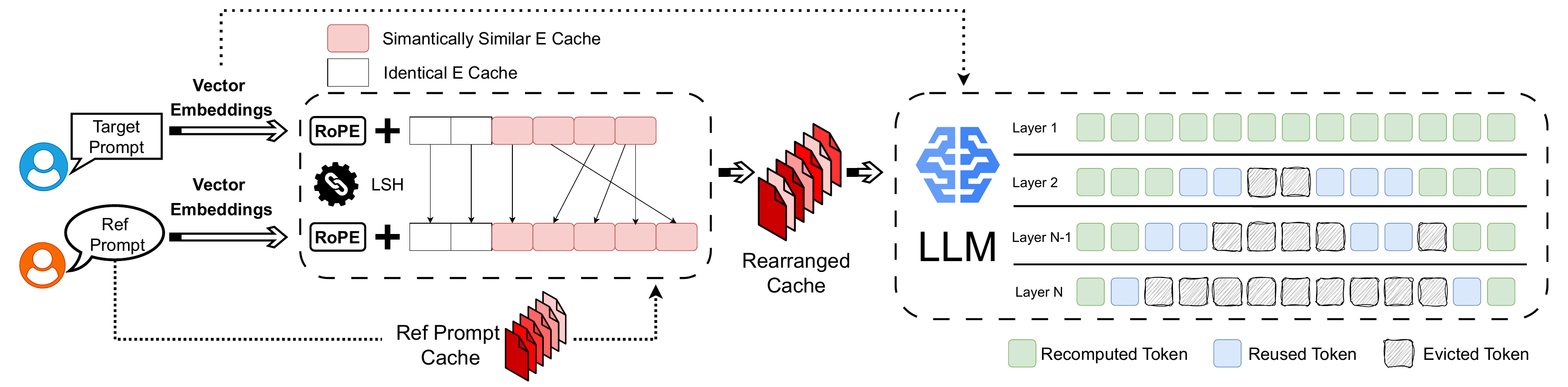}
  \caption{Schematic Overview of SemshareKV}
  \label{fig:schematic}
\end{figure*}

\section{Related Work}
Prior research on KV cache optimization can be categorized into three key directions: (i) \textbf{Conventional KVCache Compression}, which focuses on reducing the storage and computational overhead of KV cache by applying quantization, pruning, or other compression techniques; (ii) \textbf{KVCache Sharing}, which explores methods to reuse KV cache across different queries or tasks to improve efficiency while maintaining response quality; and (iii) \textbf{KVCache Reusing}, which investigates strategies to adapt and re-purpose precomputed KV cache for semantically similar inputs, minimizing redundant computation while preserving model accuracy.

\subsection{Conventional KVCache Compression} To address long-context processing, many works propose optimizing inference by retaining only informative tokens. Token-level compression often uses attention-based token selection~\citep{zhang2023h2o, xiao2024duoattention, li2024snapkv, yang2024pyramidinfer, zhong2024zigzagkv}, low-rank decomposition~\citep{sun2024shadowkv}, or quantization~\citep{zhang2024pqcache, wang2024squeezeattention}. Model-level approaches redesign architectures to improve reuse~\citep{sun2025you, yan2024recurformer}, while system-level methods focus on memory and scheduling~\citep{kwon2023efficient, sheng2023flexgen}. Recent work has highlighted the use of value vectors to facilitate compression~\citep{guo2024attention}.

\subsection{KVCache Sharing}
Some also emphasize reusing portions of the cache for future or similar queries and prompts. For example, PromptCache~\citep{gim2024prompt} stores text segments that appear frequently on an inference server using a schema, although this approach hampers usability, as users must conform their natural language to the schema format. Mooncake~\citep{qin2024mooncake}, KVSharer~\citep{yang2024kvsharer} and MiniCache~\citep{liu2405minicache} exploit the high similarity of attention scores among adjacent transformer layers to improve KV cache reuse. By consolidating or sharing KV pairs between similar layers, these methods improve memory efficiency and streamline token processing. However, their approaches are restricted to sharing in the layer or text segment within adjacent layers or the same LLM, limiting the broader applicability; GPTCache~\citep{regmi2024gpt},  \citep{rasool2024llms} and \citep{bang2023gptcache} utilize similarity search among queries to reuse KV cache. However, they have a high probability of missing a hit and require the entire query to be similar, offering limited flexibility. 

\subsection{KVCache Reusing}
Limited attention has been directed toward the sharing of KV cache in LLMs. DroidSpeak~\citep{liu2024droidspeak} improves context sharing between fine-tuned LLMs by identifying critical KV cache layers and selectively recomputing them for efficient reuse while maintaining accuracy. LMCache~\citep{cheng2024large} introduces a Knowledge Delivery Network (KDN) to optimize KV cache storage and transfer, allowing cost-effective knowledge injection in LLM inference. CacheBlend~\citep{yao2024cacheblend}, KVShare~\citep{yang2025kvshare}, and EPIC~\citep{hu2024epic} rely on exact context matching, which is unsuitable for real user scenarios. SentenceKV~\citep{Zhu2025SentenceKVEL} suffers from inter-sentence information loss, as noted in the CacheBlend. In contrast, SemShareKV introduces RoPE in token matching to address this issue.

\section{Observations and Insights}
\label{sec:three_insights}

We present three key insights derived from our experiments on three LLMs: Mistral-7B~\citep{jiang2023mistral}, LLaMA-3.1-8B~\citep{grattafiori2024llama}, and MPT-7B~\citep{MosaicML2023Introducing}. These insights show consistent patterns across different LLMs, supporting the generality of our observations.

\subsection*{Insights 1 {\textit{HD tokens stay consistent across layers.}}}

\begin{figure}[t]
  \includegraphics[width=0.9\columnwidth]{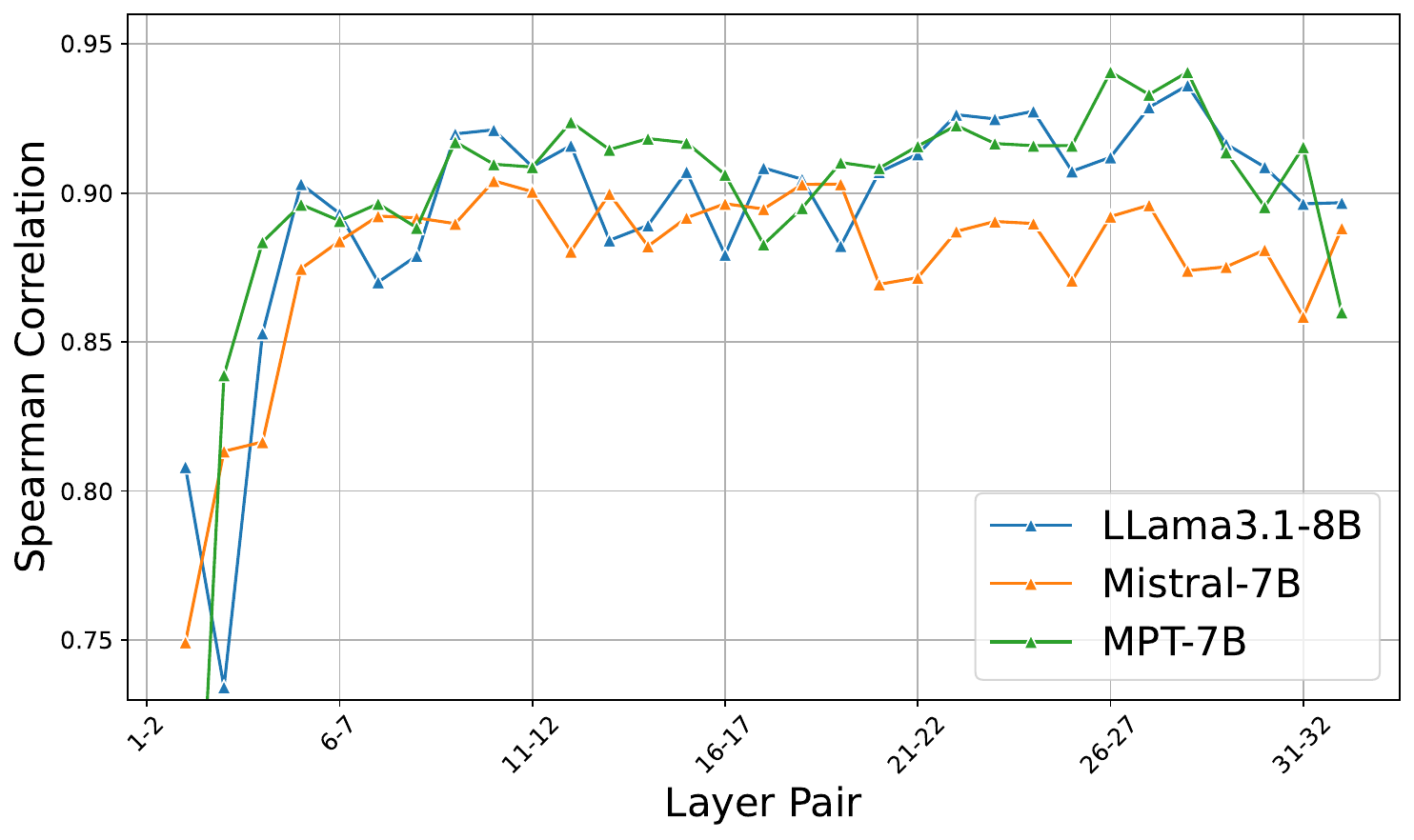}
  \caption{Insight 1: High-deviation tokens remain consistent across layers.}
  \label{fig:insight1}
\end{figure}

When reusing KV caches from semantically similar prompts, we ensure the reused cache maintains high fidelity with fully recomputed caches to prevent performance degradation. To compare the similarity between two KV matrices, we used our augmented MultiNews~\citep{bai2023longbench} dataset, where each sample consists of a pair of semantically similar prompts: the \textbf{\textit{Target Prompt}}, which serves as the primary input to the model, and the \textbf{\textit{Reference Prompt}} (Ref Prompt), which acts as the semantically similar counterpart. For each of the aforementioned LLMs, we first computed the KV caches for the prompt pairs independently. Subsequently, we calculated the deviations between the KV caches of the target and reference prompts using the previously mentioned $L_2$ norm. Tokens with the highest 40\% deviation were identified as \textbf{\textit{High Deviation}} (HD) tokens. To further quantify this observation, we computed the Spearman correlation of HD tokens between adjacent layers. As shown in Figure~\ref{fig:insight1}, adjacent layers exhibit relatively high consistency in HD token positions.

\subsection*{Insights 2 {\textit{Deeper layers focuses on fewer tokens}}}
To analyze attention patterns across layers, we first averaged the attention scores across all heads in each layer and then computed the mean along the first dimension, resulting in a one-dimensional vector per layer. To quantify this behavior, we introduce \textbf{\textit{Attention Recovery}} (AR), defined as follows:


\begin{equation}
AR = \min\{k\in[n]\mid 
  \genfrac{}{}{1pt}{}{\sum_{i=1}^k T_i}{\sum_{i=1}^n T_i} 
  \ge \text{Thres}\}
\end{equation}

\begin{figure}[t]
  \includegraphics[width=0.9\columnwidth]{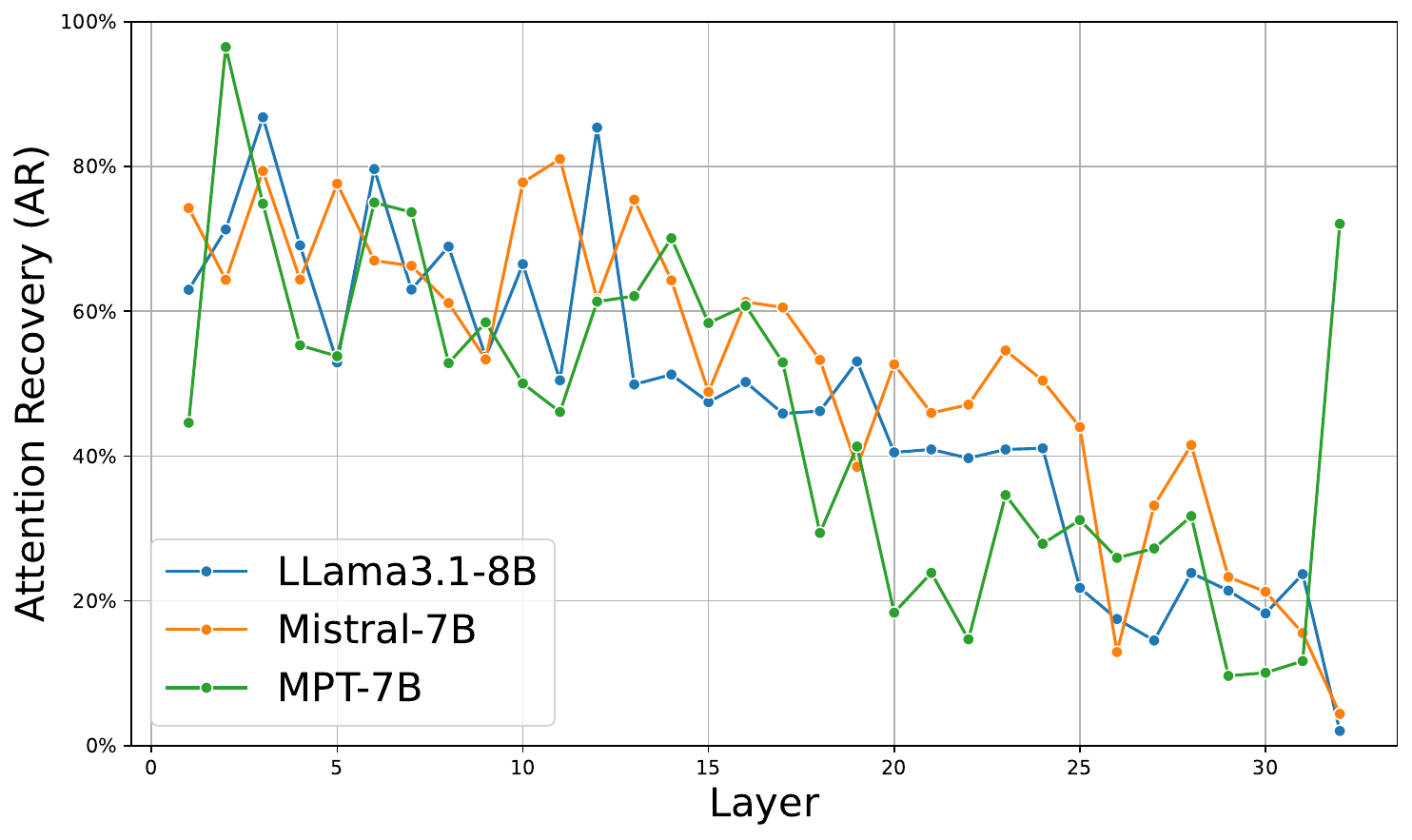}
  \caption{Insight 2: Deeper layers attend to fewer tokens.}
  \label{fig:insight2}
\end{figure}

Where $T$ is a sorted vector of average attention scores for each token, $S_{total}$ represents the total attention score derived from the averaged self-attention matrices, and $Thres$ indicates the threshold of attention score. AR indicates the number of tokens that must be summed from highest to lowest based on their average attention scores in order to cover $Thres\%$ of the total attention score. We computed AR for each layer, and the results (Figure~\ref{fig:insight2}) reveal a consistent trend: as depth increases, AR decreases across all three LLMs, despite minor fluctuations. This suggests that deeper layers concentrate attention on progressively fewer tokens, reflecting more selective focus.

\subsection*{Insights 3 {\textit{Deeper layers have more redundant information.}}}

\begin{figure*}[t]
  \centering
  \begin{subfigure}{0.32\linewidth}
    \centering
    \includegraphics[width=\linewidth]{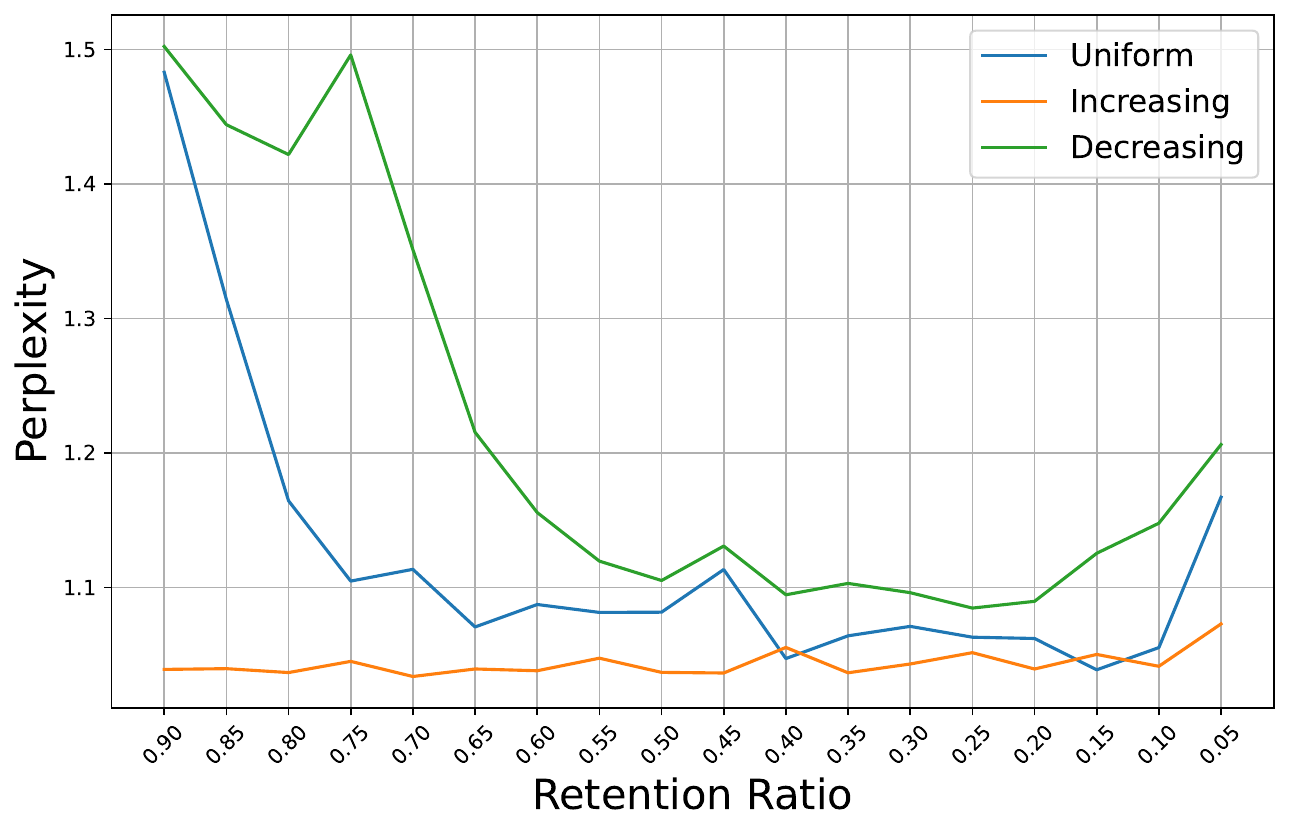}
    \caption{Llama3.1-8B Retention Pattern}
    \label{fig:insight3_1_a}
  \end{subfigure}
  \hfill
  \begin{subfigure}{0.32\linewidth}
    \centering
    \includegraphics[width=\linewidth]{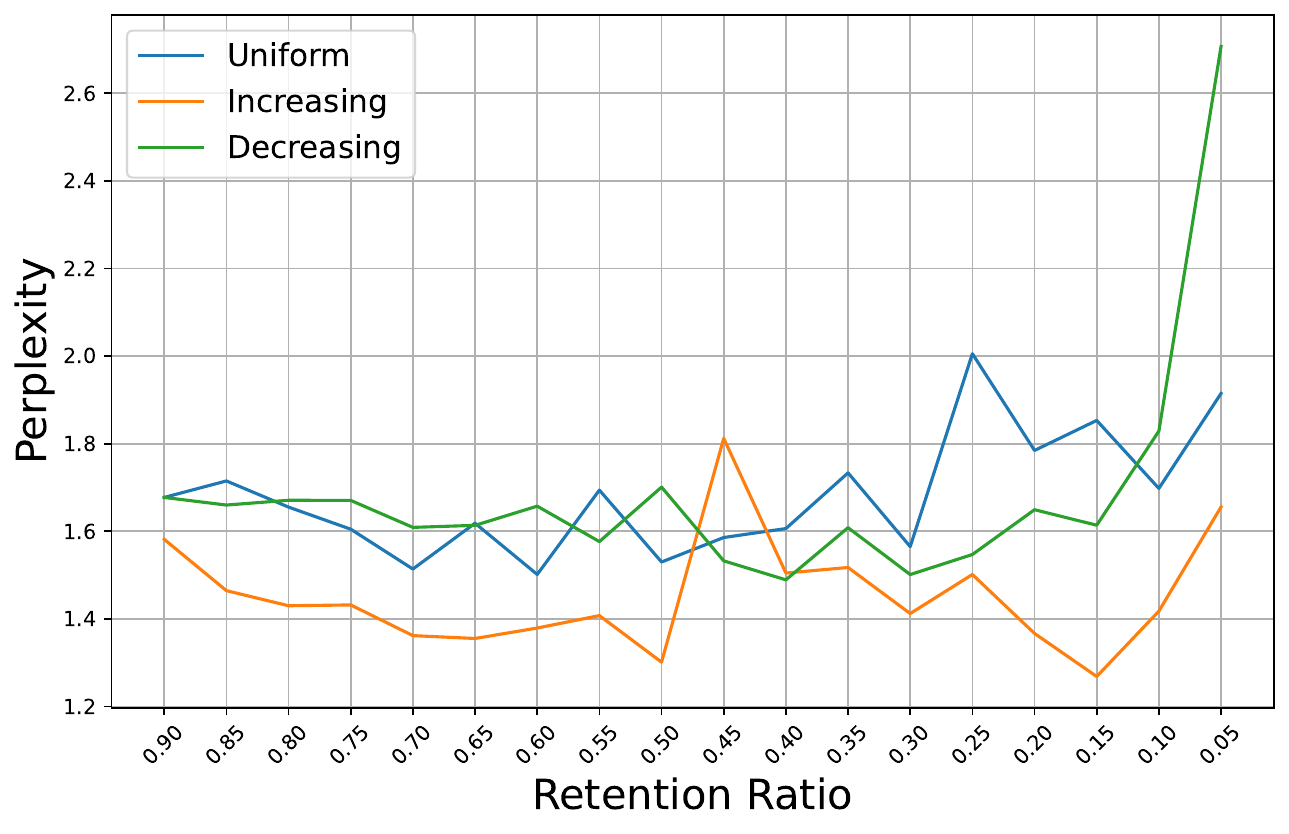}
    \caption{Mistral-7B Retention Pattern}
    \label{fig:insight3_1_b}
  \end{subfigure}
  \hfill
  \begin{subfigure}{0.32\linewidth}
    \centering
    \includegraphics[width=\linewidth]{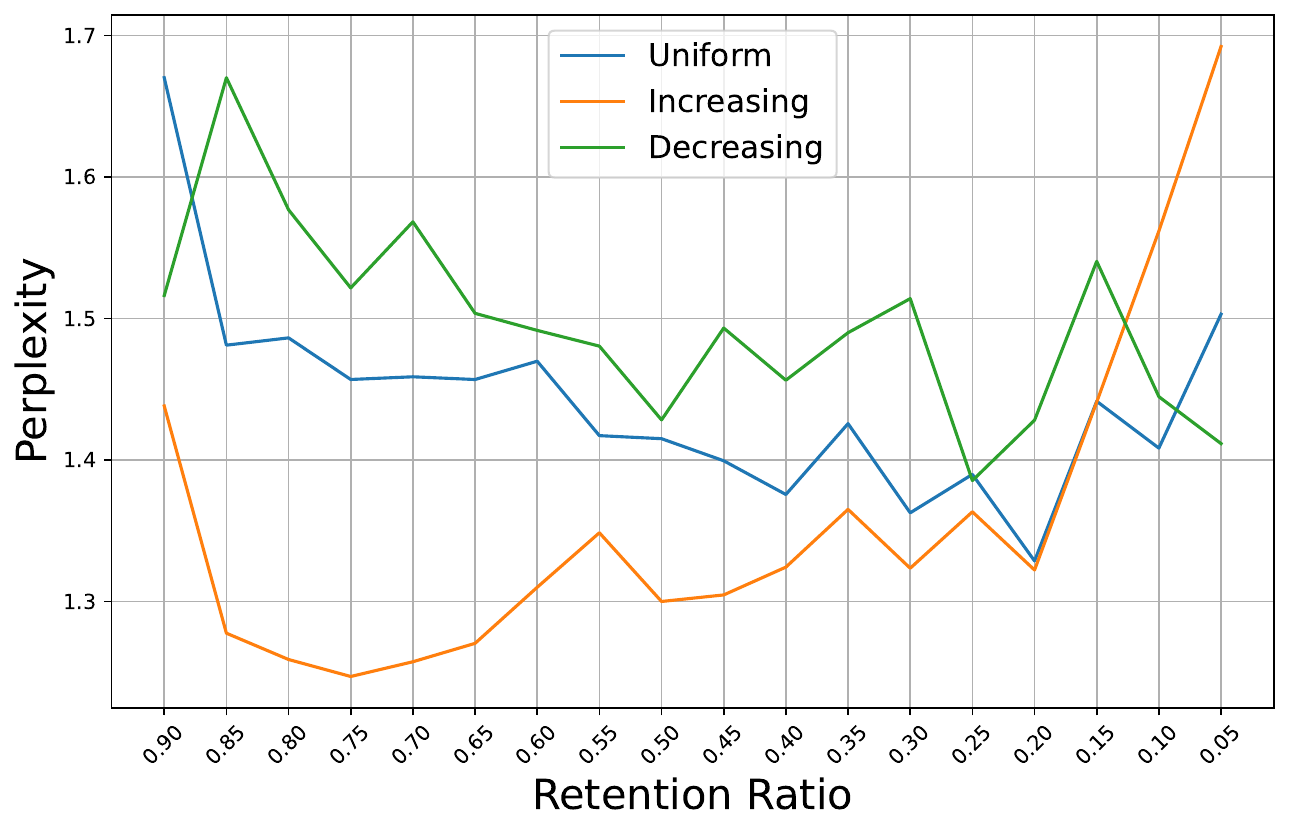}
    \caption{MPT-7B Retention Pattern}
    \label{fig:insight3_1_c}
  \end{subfigure}
  \caption{Insight 3: Deeper layers contain more redundant information.}
  \label{fig:insight3}
\end{figure*}

To reduce memory overhead from the KV cache, a key optimization strategy is to remove tokens containing redundant information. Such tokens contribute minimally to the prediction of next tokens during decoding but occupy substantial GPU memory. However, selective token retention risks information loss, necessitating careful trade-offs between memory savings and generation quality. 
We evaluate three KV cache retention strategies using perplexity: Constant, with equal retention across layers; Exponential Growth, with higher retention in shallow layers; and Exponential Decay, with more retained in deeper layers. More details of the three retention patterns can be found in Figure~\ref{fig:insight3} in the Appendix.


We applied these three retention patterns to LLMs and evaluated generation performance using perplexity. Across all three models, the Exponential Decay pattern achieves the lowest perplexity, indicating the best performance and suggesting this pattern aligns with how LLMs interpret prompt knowledge. Notably, MPT-7B exhibits a spike in AR at the final layer due to its structural differences from Mistral and Llama3.2: unlike the RoPE-based models, MPT-7B employs ALiBi positional encoding and a pre-norm design. ALiBi biases intermediate layers toward recent tokens, while the final layer compensates by attending to more distant tokens, resulting in broader attention patterns and higher AR values.

\section{Methodology}

\subsection{Relevant Concepts}
Our work focuses on three critical cache components in modern LLMs:

\begin{itemize}[leftmargin=12pt,itemsep=4pt,topsep=4pt]
    \item \textbf{Key Cache (K):} Key vectors encode the structural relationships among tokens in a sequence.
    
    \item \textbf{Value Cache (V):} Value vectors containing the actual content representations aggregated through attention weights. These preserve the contextual information of each token.
    
    \item \textbf{Embedding Cache (E):} Contextualized embeddings capturing fundamental semantic and syntactic relationships~\citep{Mikolov2013EfficientEO}, providing the fundamental token representations before transformer processing.
\end{itemize}

\subsection{Model Overview}
The design of SemShareKV, illustrated in Figure~\ref{fig:schematic}, is based on three key insights from Section~\ref{sec:three_insights}. Our approach employs two core strategies:

\begin{itemize}[leftmargin=10pt,itemsep=3pt,topsep=3pt]
    \item \textbf{Recomputation Strategy (Insights 1 \& 2):} Prioritize the recomputing of more tokens in shallow layers while reducing the recomputation in deeper layers, reflecting the varying importance of the layer depth in attention mechanisms.
    
    \item \textbf{Retention Strategy (Insights 1 \& 3):} Preserve more tokens in shallow layers while evicting tokens from deeper layers, optimizing memory usage without significant accuracy degradation.
\end{itemize}
SemShareKV stores received prompts and their corresponding contextualized E caches in CPU memory. When the LLM receives a new prompt as the target prompt, it retrieves a reference prompt by computing an LSH-distance-based similarity score between the target’s contextualized E cache and all stored E caches. The reference prompt with the highest similarity is then loaded onto the GPU along with its corresponding KV cache for reuse.

Once a reference prompt is retrieved, SemShareKV first applies RoPE to the E caches of both the target and reference prompts. Then it uses LSH to match each token in the target prompt to its most similar tokens in the reference prompt. Based on these LSH mappings, the precomputed KV cache of the reference prompt is rearranged token by token and injected into LLM transformer layers. On the first transformer layer, all tokens undergo full recomputation. The recomputed outputs are compared with the rearranged cache values via $L_2$ norm, identifying high-deviation tokens for prioritized recomputation in subsequent layers. Simultaneously, the system evicts tokens with the lowest attention scores from recent computations, optimizing KV cache memory usage dynamically.

\subsection{KVCache Sharing Challenge}
The primary challenge in cross-prompt KV cache sharing stems from length disparity between prompts. Inspired by~\citep{liu2024droidspeak}, we incorporates positional encoding within the E Cache to enable accurate token alignment while preserving contextual relationships.

Specifically, we use LSH to identify, for each token in the target prompt, the most similar token in the reference prompt based on their vector representations. We use LSH for efficient token similarity search.  This process allowed us to reorder the KV cache of the reference prompt to align with the token sequence of the target prompt. Consequently, the reordered KV cache matches the target prompt's length, with its key-value pairs entirely derived from the original KV cache of the reference prompt. The LLM uses the reordered KV cache to the target prompt. Additional details on LSH are provided in Appendix~\ref{app:lsh_extend}.

\subsubsection{Use Relative Position Encoding to Facilitate Fuzzy Token Match}

A fundamental limitation of naive matching using the E cache arises from the absence of positional context in its representation. Since raw vector embeddings lack inherent positional information, LSH fails to maintain crucial sequential relationships when identifying reference-target token correspondences. This positional agnosticism in the E cache consequently produces semantically inferior mapping results.

To address this, we introduce positional encoding into the E cache to enhance fuzzy token matching. Two widely used positional encoding strategies are absolute positional encoding~\citep{Vaswani2017AttentionIA}, which embeds explicit position information, and relative positional encoding~\citep{su2024roformer}, which captures positional relationships between tokens. In our work, we incorporate Rotary Position Embedding (RoPE) into the E cache and evaluate its impact. Specifically, RoPE is applied to the non-contextual embeddings (E cache) of both the reference and target prompts. Then, LSH is used to match each token in the target prompt’s E cache with the most similar token in the reference prompt’s E cache. This step is crucial because RoPE introduces position-sensitive information, allowing the same token at different positions to carry distinct semantics, enabling LSH to achieve more accurate token-level matching.

\begin{figure}[t]
  \centering
  \begin{subfigure}{0.48\linewidth}
    \centering
    \includegraphics[width=\linewidth]{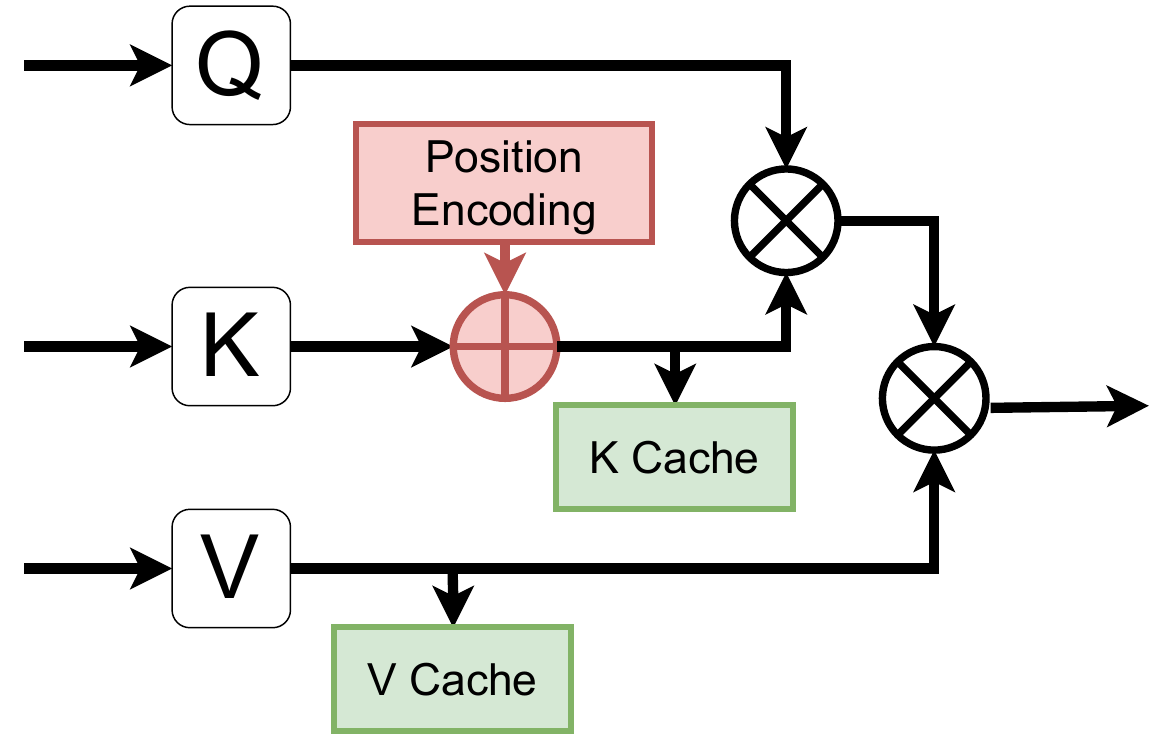}
    \caption{Standard KVCache Store}
    \label{fig:disentangle_pos_kvcache_1}
  \end{subfigure}
  \hfill
  \begin{subfigure}{0.48\linewidth}
    \centering
    \includegraphics[width=\linewidth]{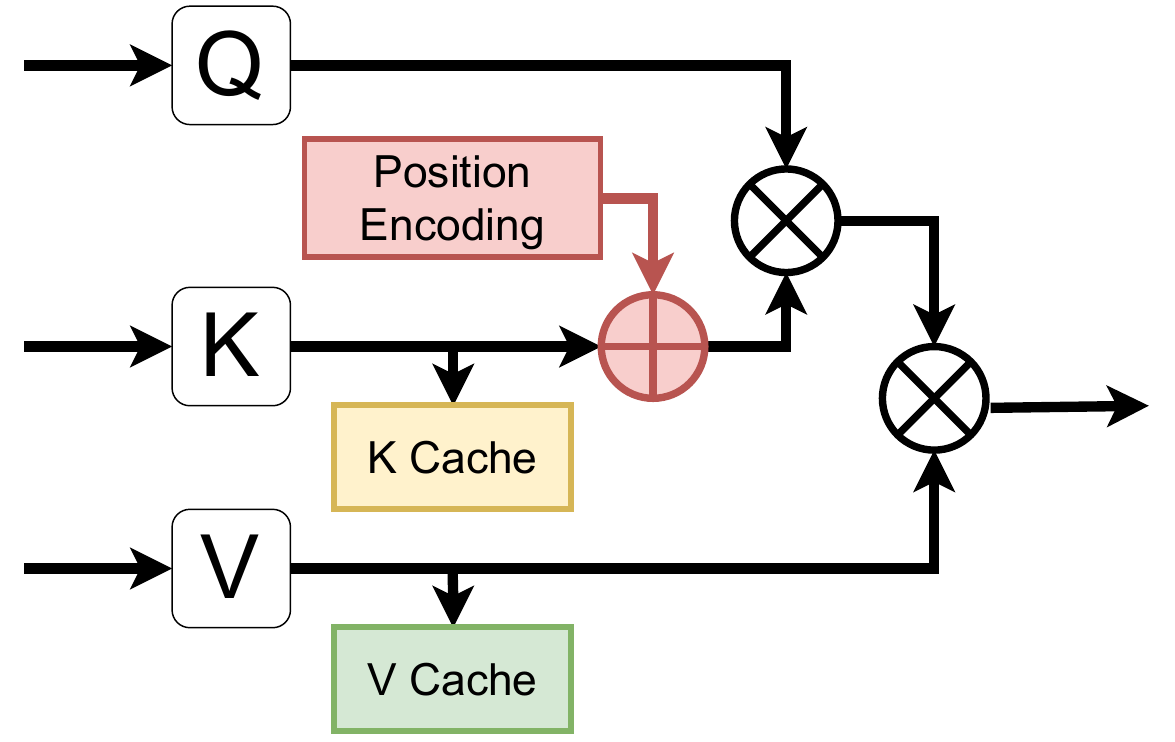}
    \caption{Modified KVCache Store}
    \label{fig:disentangle_pos_kvcache_2}
  \end{subfigure}
  \caption{Default vs. modified KV cache storage.}
  \label{fig:disentangle_pos_kvcache}
\end{figure}

\begin{figure}[t]
  \centering
  \begin{subfigure}{0.9\linewidth}
    \centering
    \includegraphics[width=\linewidth]{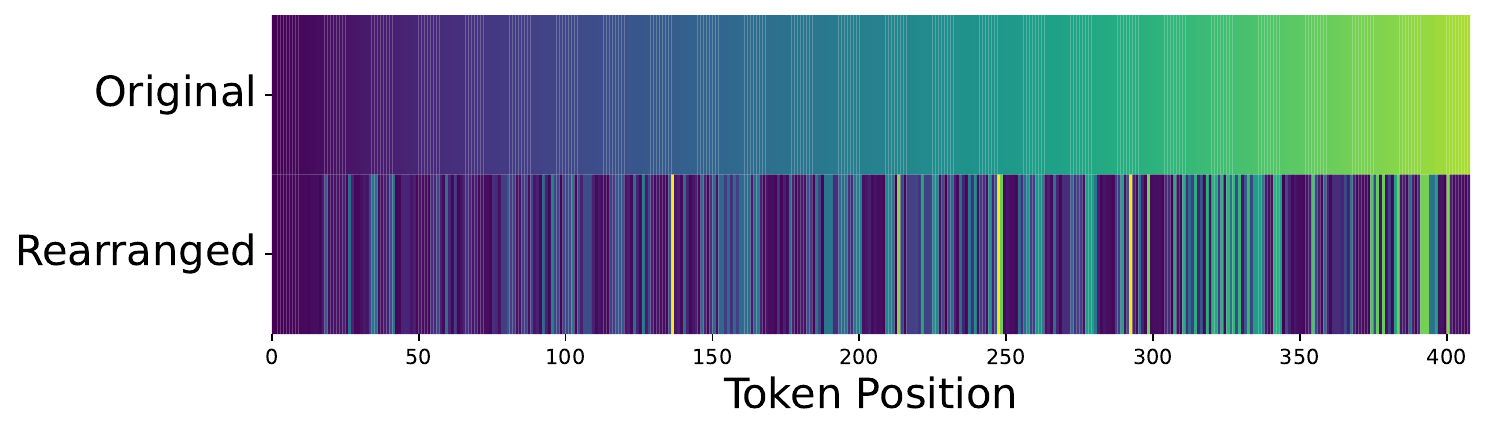}
        \caption{Only E Cache}
    \label{fig:heatmap_show_match_1}
  \end{subfigure}
  \vspace{0.2cm} 
  \begin{subfigure}{0.9\linewidth}
    \centering
    \includegraphics[width=\linewidth]{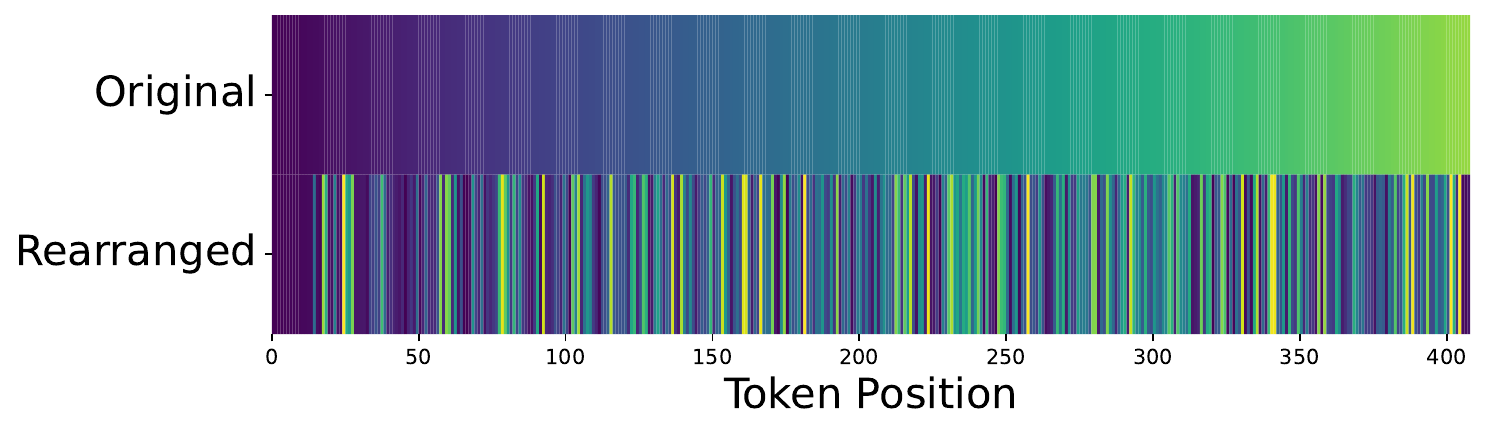}
        \caption{E Cache with Relative Position Encoding} 
    \label{fig:heatmap_show_match_2}
  \end{subfigure}
  
  \caption{Fuzzy matching: with vs. without position encoding.}
  \label{fig:encode_rope_in_e_cache}
\end{figure}

\begin{figure}[t]
  \includegraphics[width=0.9\columnwidth]{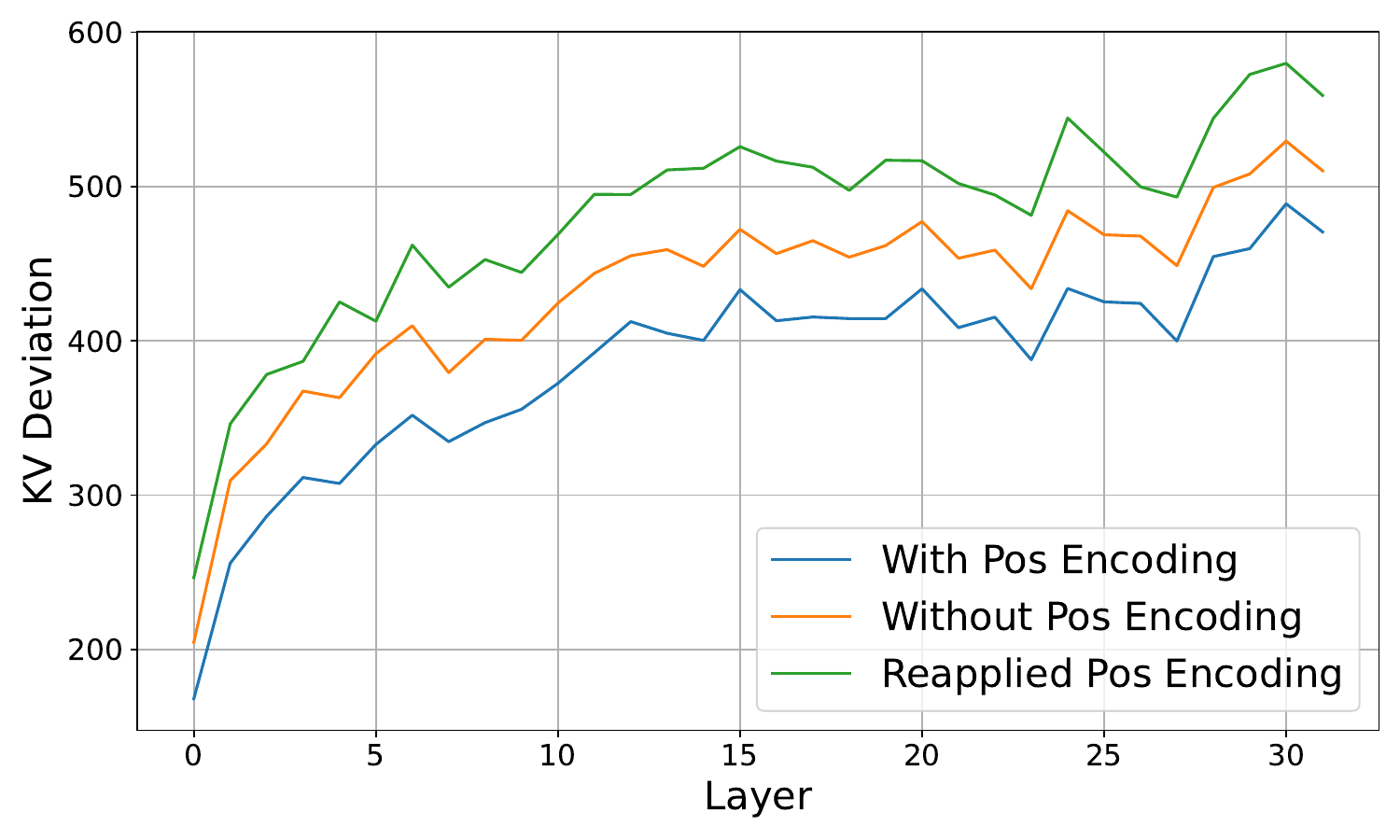}
  \caption{Impact of position encoding on E cache and KV cache deviation.}
  \label{fig:kv_dev}
\end{figure}


Figure~\ref{fig:encode_rope_in_e_cache} further illustrates how maintaining positional relationships through RoPE improves alignment accuracy, leading to better token retrieval and overall performance. Figure~\ref{fig:heatmap_show_match_1} compares the token positions in the original E cache with the rearranged positions after passing through the fuzzy token matching block, while Figure~\ref{fig:heatmap_show_match_2} presents the results when using an E cache with positional encoding. Notably, the first ~20 tokens remain in their original positions, as they represent query tokens. This demonstrates that LSH correctly identifies and preserves query positions.

Beyond the initial tokens, a key difference emerges: without positional encoding, many tokens in the target prompt map to the initial tokens, whereas with positional encoding, they align more accurately with later tokens. We interpret this as a manifestation of "\textit{attention sink}", a phenomenon in self-attention mechanisms where a significant portion of attention scores is consistently assigned to initial tokens, regardless of their actual relevance to the task~\citep{xiao2023efficient, fei2025efficient}. Incorporating positional encoding into the E cache effectively mitigates this issue, leading to more accurate token matching and improved performance.

\subsubsection{{{Impact from Position Encoding in KVCache}}}

The second challenge is that LSH-based token rearrangement disrupts position encoding in the precomputed KV cache, affecting the KV matrices from the prefill phase. Previous studies~\citep{Gao2024CostEfficientLL} have discussed the impact of different position encoding strategies, suggesting that RoPE can be excluded from KV cache by applying it after storage, as illustrated in Figure~\ref{fig:disentangle_pos_kvcache}. In evaluating the role of RoPE in the E cache, we compared three configurations of KV cache: (i) \textbf{With position encoding}, the standard setting where RoPE is applied before storing KV cache; (ii) \textbf{Without position encoding}, where RoPE is not applied during storage; and (iii) \textbf{Wthout position encoding but reapplied}, where RoPE is omitted during storage but reapplied after LSH-based reordering. Ideally, the rearranged KV cache should closely match the ground-truth KV cache for the target prompt. To quantify the deviation, we compute the \(L_2\) norm between the rearranged and ground-truth caches. As shown in Figure~\ref{fig:kv_dev}, KV cache with position encoding has the lowest deviation, followed by the version without position encoding, while the configuration with reapplied RoPE gives the highest deviation. This highlights the importance of storing KV pairs after RoPE is applied. Ensuring consistent position encoding between the E and KV cache is essential for LLMs to fully leverage them and achieve optimal generation quality.

\begin{table*}[t]
  \centering
  \small
  \setlength{\tabcolsep}{4pt}
  \renewcommand{\arraystretch}{1.15}
\caption{Performance comparison between SemShareKV and baseline methods}
  \begin{tabular}{lccccccccc}
    \hline
    \textbf{Method} & 
    \rotatebox{40}{\textbf{MultiNews}} & 
    \rotatebox{40}{\textbf{Wikihow}} & 
    \rotatebox{40}{\textbf{Qasper}} &
    \rotatebox{40}{\textbf{SAMSum}} & 
    \rotatebox{40}{\textbf{PubMed}} & 
    \rotatebox{40}{\textbf{BookSum}} & 
    \rotatebox{40}{\textbf{BigPatent}} & 
    \rotatebox{40}{\textbf{LCC}} &
    \rotatebox{40}{\textbf{MMLU}} \\
    \hline

    \multicolumn{10}{c}{\textbf{MISTRAL-7B}} \\
    \hline
    Full KV         & 22.10 & 20.50 & \textbf{17.10} & 18.79 & 24.66 & 22.44 & 25.47 & 22.41 & 34.00 \\
    SemShareKV      & 23.15 & 19.38 & 16.52 & \textbf{21.22} & 24.30 & 22.50 & \textbf{26.62} & 21.55 & 32.50 \\
    SnapKV          & 23.07 & 21.32 & 15.55 & 20.16 & 24.58 & 23.22 & 25.78 & \textbf{25.98} & \textbf{35.50} \\
    PyramidKV       & \textbf{23.71} & 20.06 & 16.34 & 20.59 & \textbf{25.13} & \textbf{23.87} & 26.20 & 15.83 & 33.00 \\
    H2O             & 23.04 & \textbf{21.33} & 15.88 & 20.50 & 23.53 & 22.77 & 24.99 & 16.57 & 33.00 \\
    \hline

    \multicolumn{10}{c}{\textbf{LLAMA3.1-8B}} \\
    \hline
    Full KV         & 22.49 & 19.71 & 14.21 & 16.69 & 24.50 & 22.65 & 27.26 & 19.01 & \textbf{55.00} \\
    SemShareKV      & 23.18 & 20.41 & 14.41 & \textbf{18.61} & 24.04 & 21.66 & 26.71 & \textbf{21.39} & 51.00 \\
    SnapKV          & \textbf{23.84} & \textbf{21.65} & 14.70 & 16.07 & \textbf{24.82} & \textbf{22.76} & 27.48 & 19.16 & 52.50 \\
    PyramidKV       & 23.71 & 20.69 & \textbf{15.56} & 16.86 & 24.50 & 22.72 & \textbf{27.53} & 14.16 & 52.00 \\
    H2O             & 22.81 & 20.61 & 14.44 & 16.81 & 24.19 & 22.08 & 26.94 & 21.32 & 47.50 \\
    \hline

  \end{tabular}

  \label{tab:benchmark_result}
\end{table*}

\subsubsection{Recomputation Strategy}
We divide the layers into two groups: the first layer and subsequent layers. Given that LLMs tend to focus more on later tokens \citep{liu2024lost, yang2024pyramidinfer}, we categorize tokens into {Cold} ($c$) and {Hot} ($h$) using a dynamic ratio $\boldsymbol{r_{dynamic}}$ from Attention Recovery with a threshold of 55\%, meaning $\boldsymbol{r_{dynamic}}\%$ of tokens with the highest cumulative attention are selected as {Hot}, and the rest as {Cold}. The total number of recomputed tokens is defined as $\mbox{Recomputed} = \omega_{c} \cdot c + \omega_{h} \cdot h$, where $\boldsymbol{\omega_{c}} = 0.1$ and $\boldsymbol{\omega_{h}} = 0.5$ in the SemShareKV.

Starting from the second layer, token selection follows this rule: based on \textit{{Insight 1}}, the tokens selected in the next layer are derived from those chosen in the previous layer based on a recompute ratio $\boldsymbol{\alpha_{recomp}}\%$ of this layer. Based on \textit{Insight 2}, $\boldsymbol{\alpha_{recomp}}\%$ in shallow layers will be relatively small while in deeper layers will be relatively large.

\subsubsection{Retention Strategy}
Similar to token recomputation, we categorize the layers into two groups: the first and the subsequent layers. On the first layer, the retention ratio is determined also by $\boldsymbol{r_{dynamic}}$, follows \mbox{${Retained} = \max(0.8, \boldsymbol{r_{dynamic}})$}. And retained tokens are selected based on average attention scores across the last $(1-\boldsymbol{r_{dynamic}})\%$ tokens, and only retain the top $\boldsymbol{r_{dynamic}}\%$ tokens with highest avg attention scores. In detail, the intuition behind selecting retained tokens is as follows: In the first layer, all hot tokens will be retained. Token eviction occurs only among Cold tokens that are not marked as recomputed. The underlying principle is that recomputed tokens provide better representations of the target prompt. If these tokens are evicted, the computational resources and time spent on recomputing them will be wasted. In subsequent layers, based on \textit{Insight 3}, we should retain fewer tokens.

\section{Evaluation and Results}
\subsection{Experiments Setup}
We select a diverse set of datasets covering a broad range of tasks. For Q\&A, we use \textbf{WikiHow}~\citep{koupaee2018wikihow} and \textbf{Qasper}~\citep{Dasigi2021ADO}. For summarization, we include \textbf{MultiNews}~\citep{bai2023longbench} (multi-document), \textbf{SAMsum}~\citep{gliwa-etal-2019-samsum} (dialogue), and \textbf{BookSum}, \textbf{PubMed}, and \textbf{BigPatent}~\citep{kwan_m4le_2023}, which represent narrative, scientific, and patent documents, respectively; all three are single-document summarization. For code completion, we use \textbf{LLC}~\citep{guo2023longcoder}. For multiple-choice Q\&A, we evaluate on \textbf{MMLU}~\citep{hendryckstest2021, hendrycks2021ethics}.

We compared SemShareKV against four baselines:  
(i) \textbf{Fully Recompute}: standard inference using the unmodified model from the Transformers library, where the entire prompt is input without any KV cache reuse; (ii) \textbf{SnapKV}~\citep{li2024snapkv}: a KV cache management method that accelerates the prefill phase by efficient caching but does not compress the KV cache; (iii) \textbf{PyramidKV}~\citep{cai2024pyramidkv}: PyramidKV leverages the Pyramidal Information Funneling pattern in LLMs by dynamically adjusting KV cache sizes across layers, allocating more in lower layers and less in higher ones; (iv) \textbf{H2O}~\citep{zhang2023h2o}: a dynamic KV cache eviction strategy that compresses KV memory by prioritizing important tokens, but does not optimize the prefill phase. 
We use a modified H2O compressing 10\% of the cache per layer, with SnapKV and H2O as baselines for prefill optimization and KV cache compression. Our framework is based on a simple LRU mechanism to manage precomputed caches, and SemShareKV can be easily integrated with other cache management strategies. The experiments were run on a single A100 GPU with standard attention. The implementation details are in Appendix~\ref{app:implementation}.

To the best of our knowledge, no existing dataset benchmarks LLMs on KV cache sharing across semantically similar prompts. To bridge this gap, we constructed evaluation samples by randomly selecting portions of entries from existing datasets and rewriting them using the Llama3 model. Then, these rewritten samples were manually verified to ensure that they remained semantically close to the originals. More details in the data preparation are provided in the Appendix~\ref{app:data_prepare}. 

\subsection{Benchmarking Evaluation}
We argue that using Fuzzy Token Match introduces only a negligible overhead to model inference. Table~\ref{tab:benchmark_result} reports the ROUGE-L scores~\citep{lin2004rouge}. Benchmarking results show that SemShareKV achieves performance comparable to or better than other baseline methods. Notably, in most of the evaluated datasets, \textit{Fully Recompute} fails to attain the highest performance scores. We attribute this phenomenon to the token eviction mechanisms employed by SemShareKV, SnapKV, PyramidKV and H2O. By selectively retaining only the most semantically significant tokens for self-attention computation, these methods effectively reduce redundant information in the semantic representation, thereby enhancing the model's generation quality.

\begin{figure}[t]
  \centering
  \begin{subfigure}{\linewidth}
    \centering
    \includegraphics[width=0.9\linewidth]{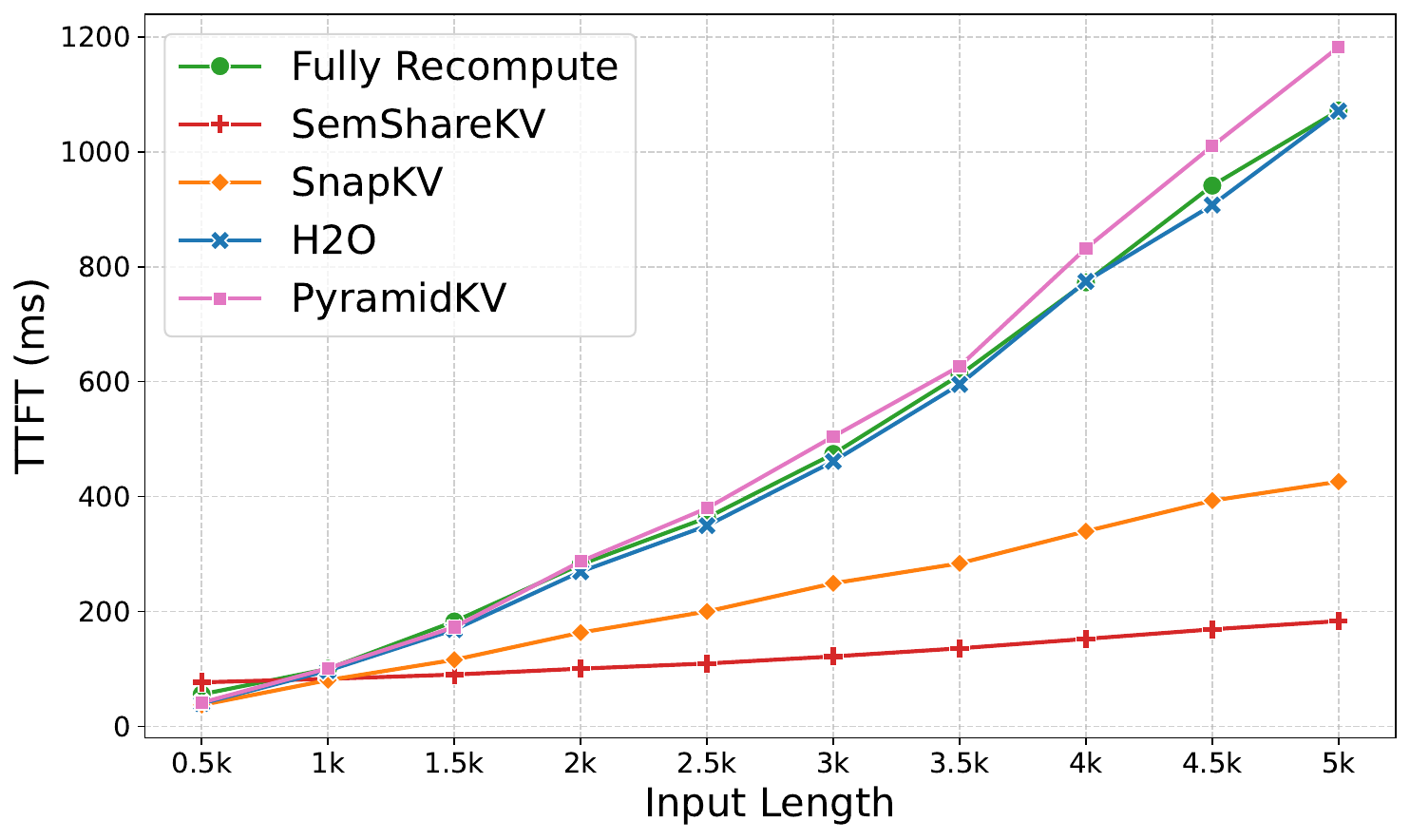}
    \caption{TTFT Comparison}
    \label{fig:effi_eval_ttft_multinews}
  \end{subfigure}
  \vspace{0.5em}
  \begin{subfigure}{\linewidth}
    \centering
    \includegraphics[width=0.9\linewidth]{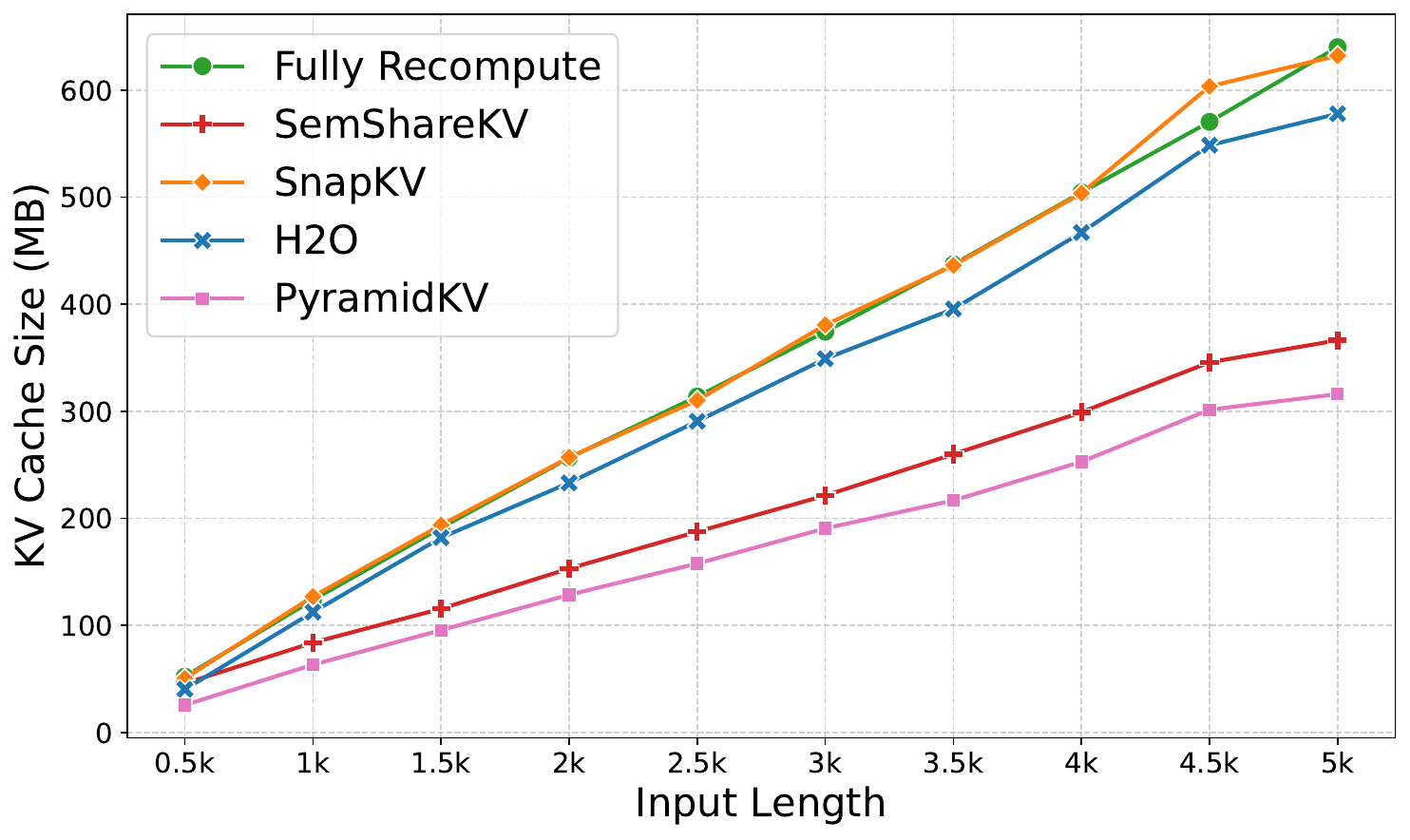}
    \caption{KVCache Size Comparison}
    \label{fig:effi_eval_kvsize_multinews}
  \end{subfigure}
  \caption{Efficiency Evaluation Results.}
  \label{fig:effi_eval_multinews}
\end{figure}


\begin{figure}[t]
  \includegraphics[width=\columnwidth]{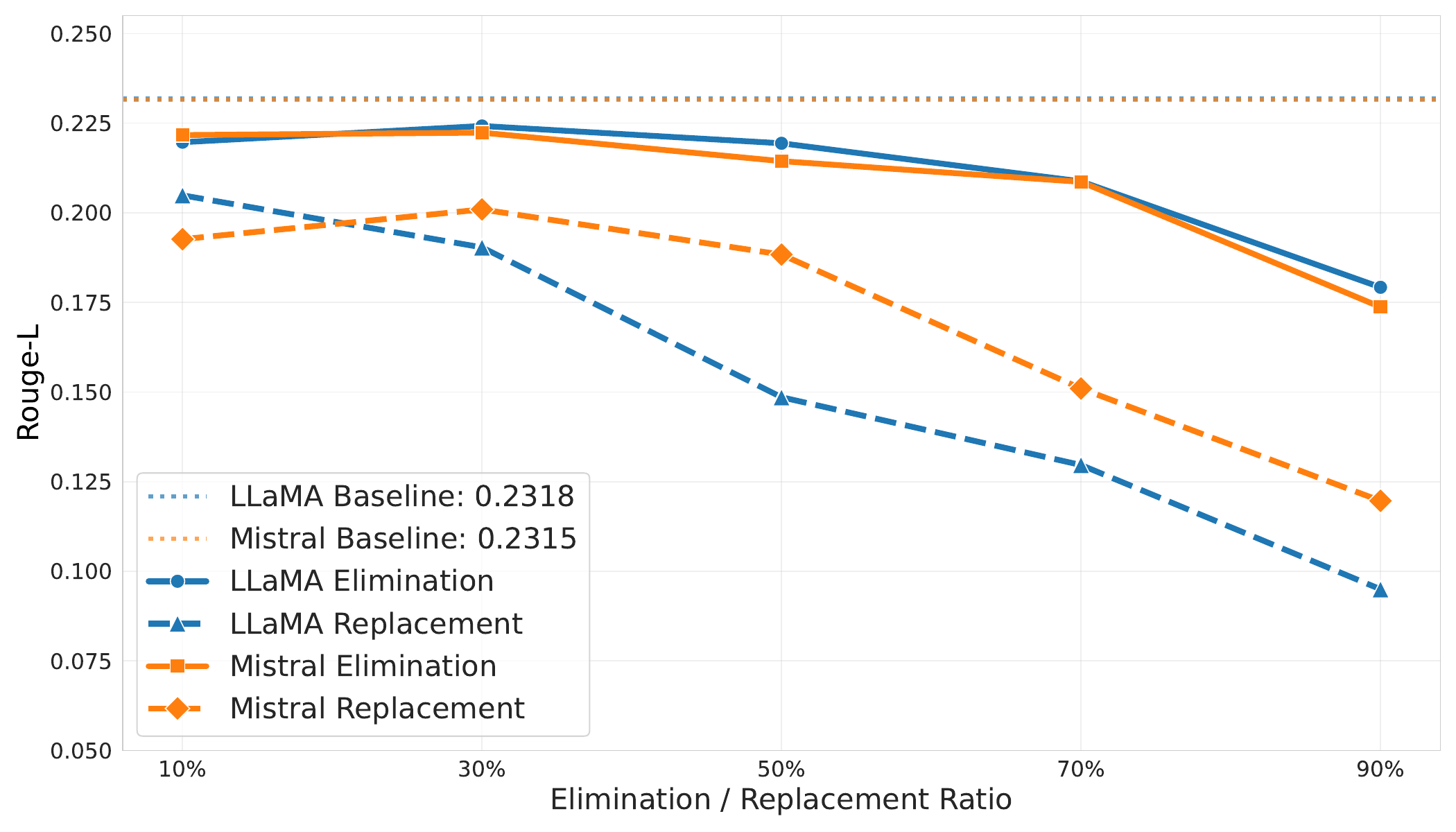}
  \caption{Impact of Prompt Similarity}
  \label{fig:elliminate_replace}
\end{figure}

\begin{figure}[t]
  \includegraphics[width=0.9\columnwidth]{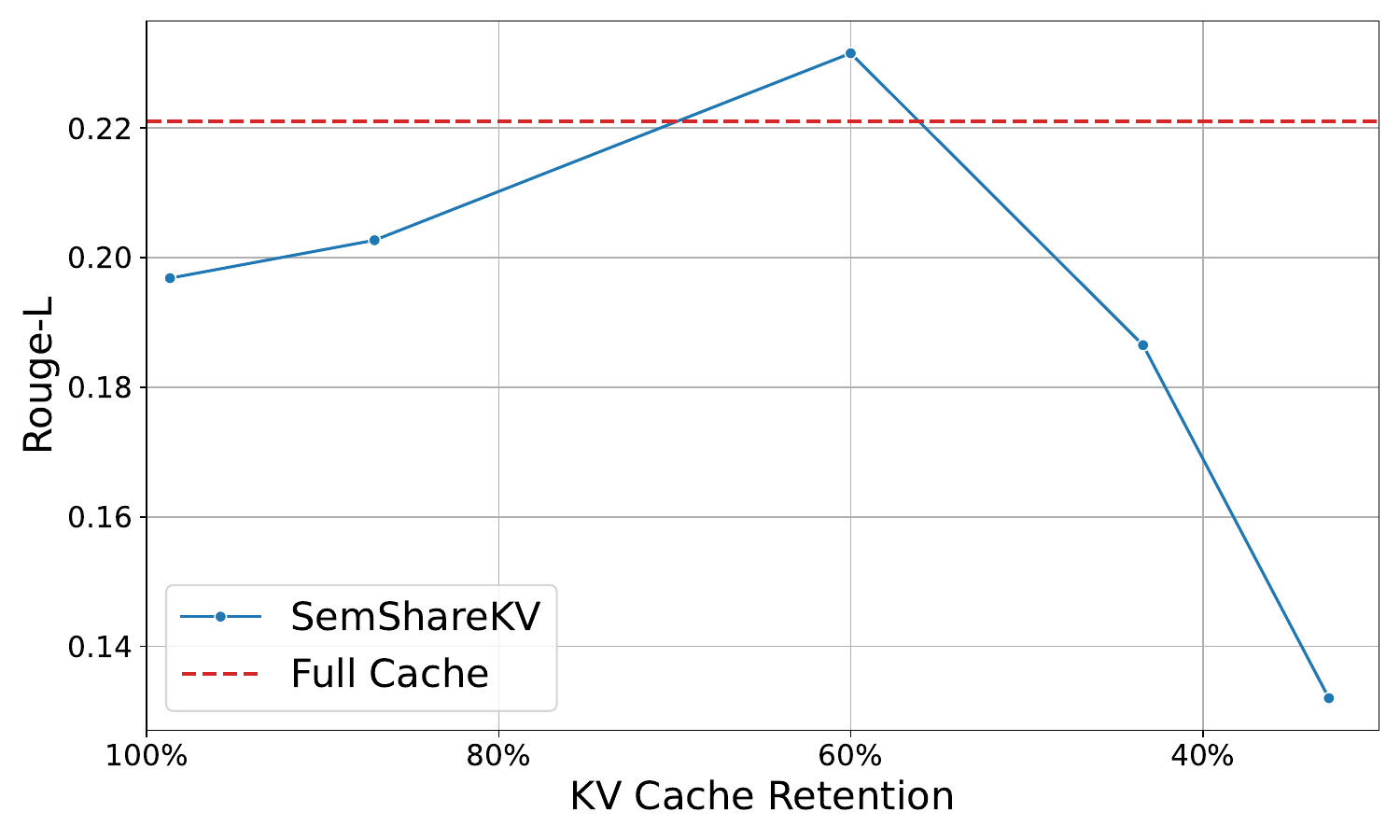}
  \caption{Cache Retention Ratio and Performance}
  \label{fig:relatio_kvcahce_rouge}
\end{figure}

\subsection{Efficiency Evaluation}
We evaluate SemShareKV based on Time To First Token (TTFT) and KV cache GPU KV memory usage, benchmarking it against Fully Recompute, SnapKV, and the unmodified H2O model. Figure~\ref{fig:effi_eval_multinews} demonstrates the efficiency advantages of SemShareKV on the MultiNews dataset, showing consistent improvements over baseline methods: SemShareKV achieves \textbf{6.25$\times$} faster Time-To-First-Token (TTFT) than Fully Recompute, PyramidKV and H2O, \textbf{2.23$\times$} faster TTFT than SnapKV, while reducing memory usage by \textbf{42\%}. Although PyramidKV uses less KV cache than SemShareKV, it does not accelerate the prefill phase, highlighting the effectiveness of SemShareKV. However, as illustrated in Figure~\ref{fig:effi_eval_kvsize_multinews}, SemShareKV offers limited performance improvements for shorter prompts (fewer than 700 tokens), which we attribute to the overhead caused by fuzzy token matching and the rearrangement of tokens from the precomputed cache of the reference prompt. In future work, our goal is to minimize this overhead.

\begin{table}[t]
  \centering
  \small
  \setlength{\tabcolsep}{5pt}
  \renewcommand{\arraystretch}{1.1}
    \caption{Ablation study on ROUGE-L for SemShareKV and its ablations across datasets.}

  \begin{tabular}{lcc}
    \hline
    \textbf{Method} & \textbf{SAMSum($\uparrow$)} & \textbf{MultiNews($\uparrow$)} \\
    \hline
    SemShareKV      & \textbf{21.22}   & \textbf{23.15} \\
    Fuzzy + Full Cache   & 17.27   & 21.34 \\
    Ablation-Zero   & 14.63   & 17.71 \\
    Ablation-Random & 5.38   & 12.67 \\
    \hline
  \end{tabular}
  \label{tab:ablation_ppl}
\end{table}

\subsection{Impact of Prompt Similarity on Performance.}
To evaluate similarity effects and SemShareKV performance, we designed two studies using the same percentage range (10\% to 90\% in 10\% increments): 1) \textbf{Randomly eliminating} sentences from the context, and 2) \textbf{Randomly replacing} sentences with others from the MultiNews dataset. We then applied SemShareKV to reuse the cache from the modified reference prompt for the target prompt. As shown in Figure~\ref{fig:elliminate_replace}, performance gradually degrades as more sentences are removed, yet remains reasonable even with substantial reductions. Notably, SemShareKV maintains strong performance even when 50\% of context sentences are removed, highlighting the effectiveness of LSH-based token-level matching. This trend holds across both LLMs, suggesting the generality of our approach. Additionally, based on the observed performance trend, we empirically set a threshold of 0.8 for applying SemShareKV, meaning that if two prompts have an LSH similarity score above 0.8, SemShareKV can be applied. Figure~\ref{fig:elliminate_replace} illustrates how the LSH-distance-based similarity changes as the replacement and elimination ratios increase. More details are in Appendix~\ref{app:ablation_study_data}.

\subsection{Ablation Study}
We conducted three ablation studies to assess the effect of fuzzy token matching on semantic understanding. First, when applying the fuzzy matching technique with the full cache, we observed a decline in performance, indicating the necessity of the token retention mechanism. Second, when matched KV cache tokens were either zeroed out or replaced with random ones, the ROUGE-L scores dropped significantly compared to the full SemShareKV method, confirming that fuzzy matching is crucial to capture semantic relationships (Table~\ref{tab:ablation_ppl}). Third, as shown in Figure~\ref{fig:relatio_kvcahce_rouge}, an analysis of the KV cache compression ratio in the MultiNews dataset revealed that retaining too much cache introduces redundancy and degrades performance, while retaining too little leads to information loss, highlighting the importance of a balanced cache retention strategy.

\section{Conclusion}
We proposed SemShareKV, a KV cache sharing framework that enables reuse across semantically similar prompts through fuzzy token matching using LSH. SemShareKV achieves a speed of 6.25$\times$ and saves up to 42\% KV cache memory space compared to conventional KV cache, with a minimum performance drop.

\section*{Limitations}
Although SemShareKV effectively shares KV caches across semantically similar prompts, the speedup decreases for shorter prompts because of the overhead introduced by fuzzy token matching, and several hyper-parameter require careful tuning. Our current implementation focuses on demonstrating the effectiveness of SemShareKV and does not yet support FlashAttention, which we plan to explore into future work. The matching threshold is also empirically determined, and exploring adaptive strategies remains an open direction. In the ablation study, we replaced sentences with other samples from the same dataset, ensuring a fair comparison within a consistent data distribution. Although cross-corpus replacement could test broader robustness, our setup effectively isolates the impact of semantic mismatches. This opens future work not only to extend the experiments to cross-corpus settings but also to construct datasets of semantically similar prompts to better evaluate robustness.


\bibliography{main}

\appendix
\section{Formula and Inference}
\label{sec:appendix}
\renewcommand{\thetable}{B\arabic{table}}
\renewcommand{\thefigure}{C\arabic{figure}}
\setcounter{table}{0}
\setcounter{figure}{0}

\begin{table*}[t]
  \centering
  \small
  \setlength{\tabcolsep}{4pt}
    \caption{Similarity evaluation on benchmarking datasets using ROUGE-L, BERTScore, and BLEU.}
  \renewcommand{\arraystretch}{1.1}
  \begin{tabular}{lcccccccccc}
    \hline
    \textbf{Metric} &
    \rotatebox{40}{\textbf{MultiNews}} & 
    \rotatebox{40}{\textbf{Wikihow}} & 
    \rotatebox{40}{\textbf{Qasper}} &
    \rotatebox{40}{\textbf{SAMSum}} & 
    \rotatebox{40}{\textbf{PubMed}} & 
    \rotatebox{40}{\textbf{BookSum}} & 
    \rotatebox{40}{\textbf{BigPatent}} & 
    \rotatebox{40}{\textbf{LCC}} &
    \rotatebox{40}{\textbf{MMLU}} \\
    \hline
    \noalign{\vskip 0.5mm}
    N of Samples    & 100 & 100 & 100 & 100 & 100 & 100 & 100 & 100 & 200 \\
    Rewrite \% (Avg)    & 54.58 & 100 & 28.99 & 46.75 & 45.64 & 44.31 & 28.55 & 29.51 & 76.39 \\
    \noalign{\vskip 0.5mm}
    \cline{1-10}
    \noalign{\vskip 1mm}
    ROUGE-L(\%) & 84.71 & 83.82 & 91.71 & 50.90 & 88.15 & 78.15 & 90.03 & 87.34 & 44.04 \\
    BERTScore(\%)    & 95.85 & 95.98 & 98.13 & 86.97 & 95.48 & 95.58 & 96.07 & 98.41 & 89.57 \\
    BLEU(\%)    & 90.40 & 87.84 & 91.32 & 24.68 & 89.22 & 81.16 & 89.29 & 89.76 & 40.51 \\
    \hline
  \end{tabular}

  \label{tab:data_prepare}
\end{table*}

\begin{figure*}[t]
  \centering
  \begin{subfigure}{0.49\linewidth}
    \centering
    \includegraphics[width=\linewidth]{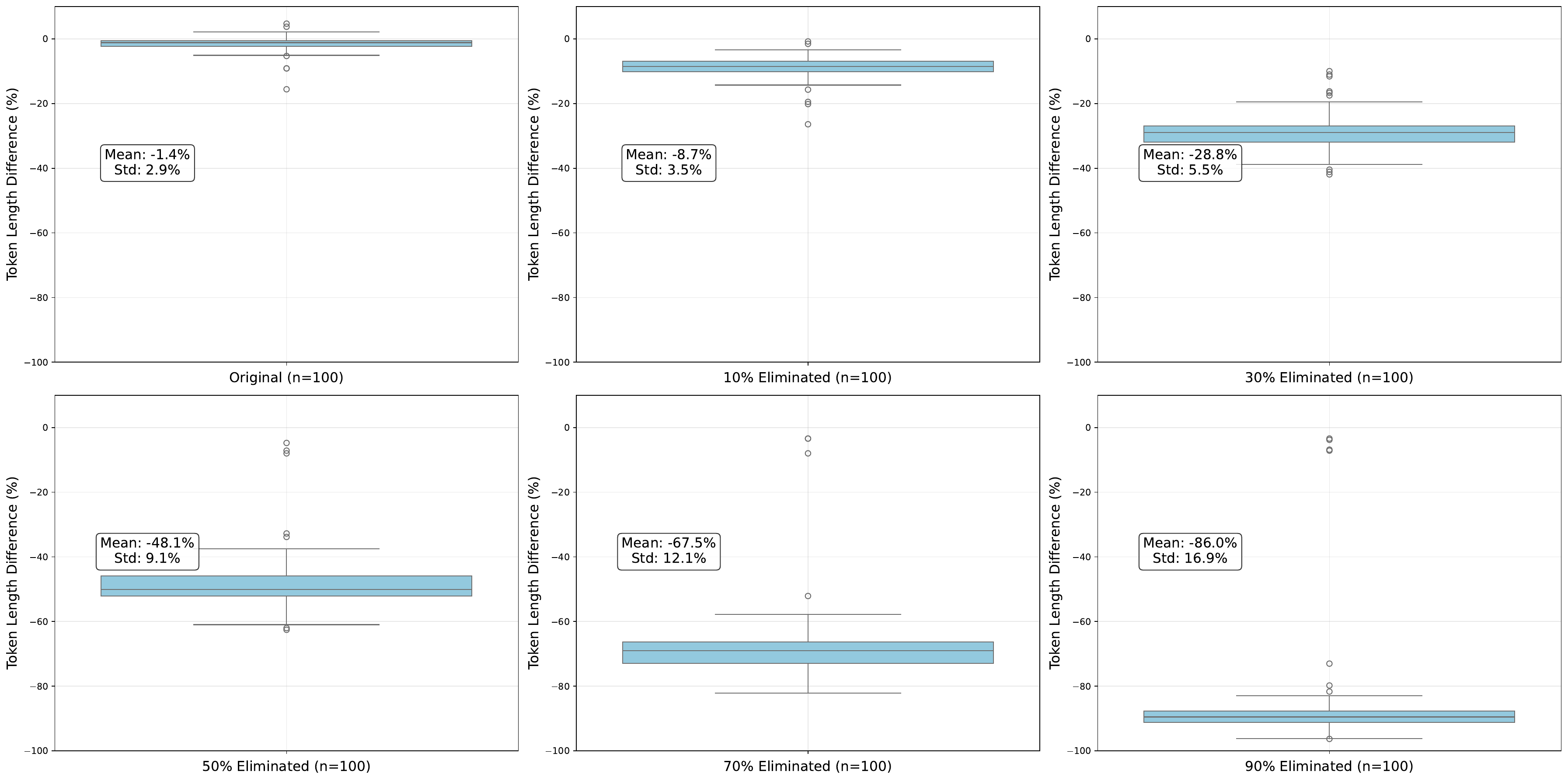}
  \caption{Elimination from Multinews dataset}
    \label{fig:elimi_replace1_a}
  \end{subfigure}
  \hfill
  \begin{subfigure}{0.49\linewidth}
    \centering
    \includegraphics[width=\linewidth]{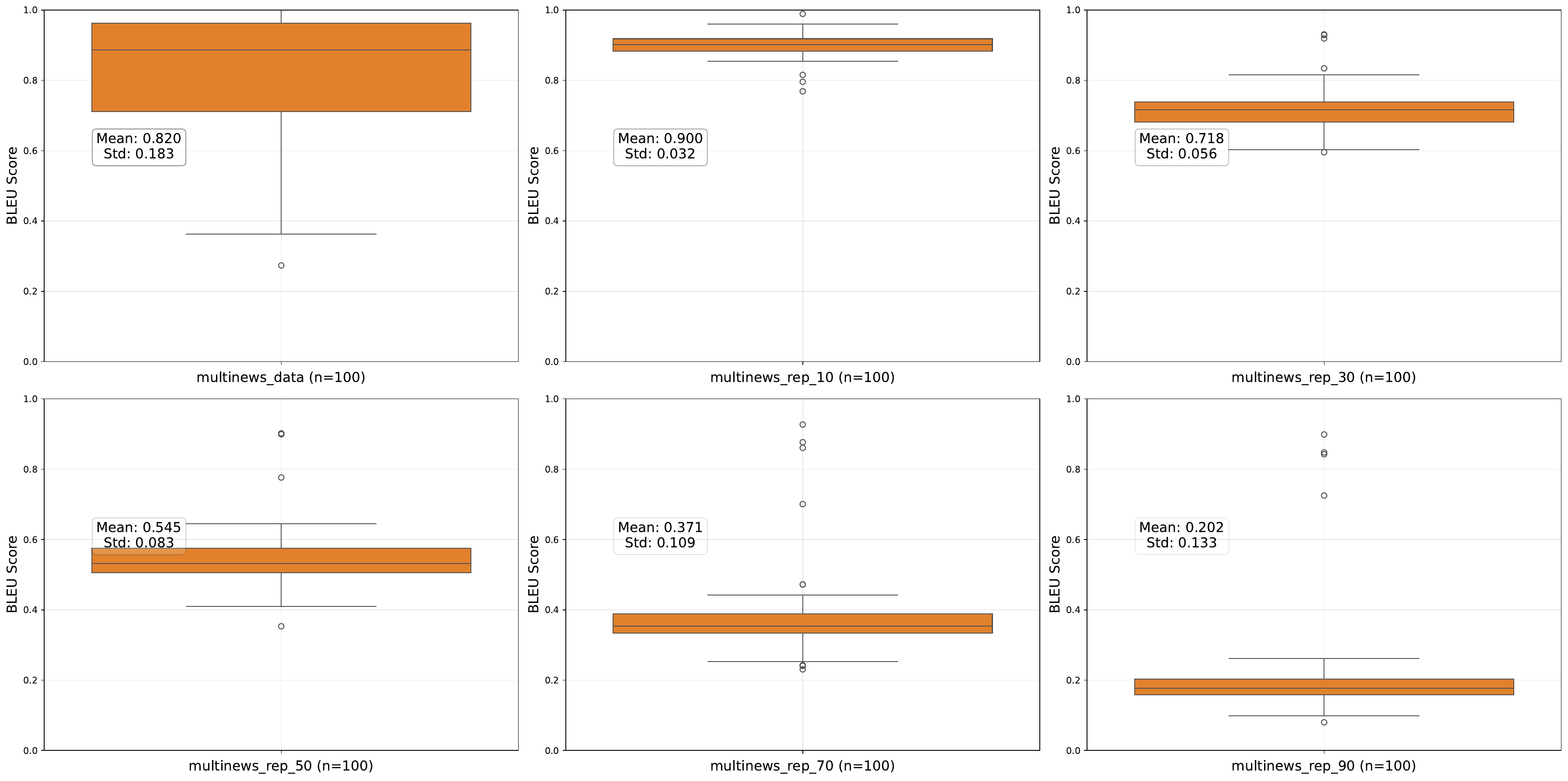}
  \caption{Replace from Multinews dataset}
    \label{fig:elimi_replace1_b}
  \end{subfigure}
  \caption{Insight 3: Deeper layers contain more redundant information.}
  \label{fig:elimi_replace}
\end{figure*}

\subsection{Rotary Position Encoding}
In our methodology, we introduced the application of \textbf{Rotary Position Embedding (RoPE)} to the E cache, which improves the performance of fuzzy token matching. RoPE is designed to incorporate positional information directly into embeddings, allowing for improved alignment between tokens in a sequence. This is particularly important in natural language processing tasks where the order of words can significantly impact the meaning and context.

The formula of RoPE in a 2-D case is shown below:

\begin{equation}
\text{RoPE}(\mathbf{x}) =
\begin{bmatrix}
\cos(\theta_k) & -\sin(\theta_k) \\
\sin(\theta_k) & \cos(\theta_k)
\end{bmatrix}
\begin{bmatrix}
x_{2k} \\
x_{2k+1}
\end{bmatrix}
\tag{1}
\label{eq:rope}
\end{equation}
In this equation, \(\theta_k = 10000^{-2k/d}\), where \(d\) represents the embedding dimension. The use of RoPE allows for the effective encoding of relative positional information, enabling the model to better capture the relationships between tokens in a sequence. Integrating RoPE into the E cache facilitates the identification of semantically similar tokens using LSH, leading to more accurate and efficient fuzzy token matching. This enhancement helps the model perform more accurately on tasks that require strong semantic understanding. 

\subsection{Locality-Sensitive Hashing (LSH)}
\label{app:lsh_extend}
Locality-Sensitive Hashing (LSH) is a technique that enables efficient approximate nearest neighbor searches in high-dimensional spaces by ensuring similar input items are hashed into the same bucket with high probability~\citep{indyk1998approximate}. This reduces the number of distance computations required, making LSH particularly useful for large datasets in applications such as image retrieval and natural language processing~\citep{datar2004locality}.
In LSH for Euclidean distance, a common hash function is:

\[
h(x) = \lfloor \frac{x \cdot r + b}{w} \rfloor
\]

where \( r \) is a random vector, \( b \) is a random offset, and \( w \) is the hash width. This overview encapsulates the theory and practical application of LSH in our framework.

\subsection{LSH-Distance Based Similarity Score}
For retrieving reference prompts to reuse cache with \textsc{SemShareKV}, we compute a similarity score by normalizing the LSH distance and inverting it to fit within a [0, 1] range:
\begin{equation}
\begin{aligned}
d_{\text{norm}} &= \frac{\mathrm{LSH\_dist} - \min(\mathrm{dist})}
{\max(\mathrm{dist}) - \min(\mathrm{dist})} \\
\text{Similarity} &= \operatorname{clip}(1 - d_{\text{norm}},\, 0,\, 1)
\end{aligned}
\end{equation}
where \( d_{\mathrm{norm}} \) denotes the normalized LSH distance; $\min(\mathrm{dist})$ is set to 0 and $\max(\mathrm{dist})$ is set to 30.

\subsection{Key-Value Deviation}

We define Key-Value Deviation with $L_2$ norm as below: 

\begin{equation}
  \begin{aligned}
    \sigma_{K} &= \lVert K^{\text{reused}} - K^{\text{recomputed}} \rVert_2, \\
    \sigma_{V} &= \lVert V^{\text{reused}} - V^{\text{recomputed}} \rVert_2, \\
    \sigma_{KV} &= {\sigma_{K} + \sigma_{V}}
  \end{aligned}
  \label{eq:sigma_kv}
\end{equation}
Where $K^{\text{reused}}$ and $V^{\text{reused}}$ represent the Key and Value matrices in cache reused from the semantic similar prompt;  $K^{\text{recomputed}}$ and $V^{\text{recomputed}}$ refer to the Key and Value matrices recomputed at the current layer.

\subsection{Token Recomputation}
The total number of tokens recomputed on layer \textit{i} is represented as 
\begin{equation}
    \begin{aligned}
        \text{Recomp}[i] = \text{T}\prod_{j=1}^{i} \alpha_{\text{recomp}}[j]
    \end{aligned}
\end{equation}
Where \( T \) denotes the total number of tokens, $\boldsymbol{i}$ represents the layer index.

\subsection{Token Retention}
The token retained on each layer is defined as:
\begin{equation}
    \begin{aligned}
        \text{Retain}[i] = \text{T}\prod_{j=1}^{i} \alpha_{\text{retain}}[j]
    \end{aligned}
\end{equation}
Where \( T \) is the total tokens, \( \boldsymbol{i} \) the layer index, and \( \alpha_{\text{retain}}[j] \) the token retention ratio at layer \( j \); tokens not retained are evicted. Typically, \( \alpha_{\text{retain}} \) is larger in shallow layers and smaller in deeper ones.

\section{Data Preparation}
\label{app:data_prepare}
\subsection{Benchmark Datasets}
We categorize these nine English-language datasets into four groups based on how semantically similar samples are constructed and the nature of the task.

\begin{enumerate}
    \item \textbf{MultiNews~\citep{bai2023longbench}:} This datasets contain samples composed of multiple independent passages or articles. To generate semantically similar samples, we randomly select one passage or article from each sample and use the Llama 3 model to rewrite it while preserving the original semantics. The rewritten passage is constrained to have a similar length to the original (within a 10\% difference in token count). We then replace the original passage with the rewritten one to construct a semantically similar prompt. The position of the rewritten passage naturally varies across samples, appearing at the beginning, middle, or end of the context.

    \item \textbf{SAMSum~\citep{gliwa-etal-2019-samsum}, PubMed, BigPatent, BookSum~\citep{kwan_m4le_2023}, LCC~\citep{guo2023longcoder}:} These datasets consist of semantically continuous text or codes. For each sample, we divide the context into individual sentences and randomly select a contiguous segment of the total sentence count. This segment is rewritten using the Llama 3 model, with the constraint that the token count deviates by less than 10\% from the original. The rewritten segment replaces the original to create a semantically similar prompt, with its position varying within the context in a similar manner.

    \item \textbf{Qasper~\citep{Dasigi2021ADO} and WikiHow~\citep{koupaee2018wikihow}:} These datasets consist of Q\&A tasks where each question must be answered based on a specific provided context. To preserve the accuracy of the questions, we use the LLM to rewrite only part of the context, leaving the questions unchanged.

    \item \textbf{MMLU~\citep{hendrycks2021ethics}:} MMLU is a multiple-choice question-answering dataset. To ensure the logical integrity of the questions and preserve the original answers, we prompt the LLM to paraphrase each entire question.
\end{enumerate}

Table~\ref{tab:data_prepare} presents the results of the similarity evaluation, measured using ROUGE-L~\citep{lin2004rouge}, BLEU~\citep{papineni2002bleu}, and BERTScore~\citep{zhang2019bertscore}. We include both longest common subsequence-based metrics (ROUGE-L), n-gram-based metrics (BLEU) and embedding-based metrics (BERTScore) to provide a comprehensive evaluation of semantic similarity across rewritten datasets.

\subsection{Eliminination and Replacement Dataset}
\label{app:eliminate_data_lsh_simi}


To study the impact of prompt similarity on LLM performance when applying SemShareKV, we designed two ablation studies. In the first, we randomly removed a portion of sentences from each sample in the \textsc{MultiNews} dataset, then applied SemShareKV to evaluate its effectiveness. Figure~\ref{fig:elimi_replace1_a} presents box plots of token length differences in the Elimination datasets compared to the original dataset. Figure~\ref{fig:elimi_replace1_b} shows the BLEU scores of the Replacement datasets relative to the original dataset.

\label{app:ablation_study_data}

\section{Extra Experimental Results}

Figure~\ref{fig:insight3} illustrates the three retention patterns discussed in \textbf{Insights 3}. Table~\ref{tab:throughput_compare_table} provides additional benchmarking results of Total Batch Time (TBT) and throughput. Since all experiments were conducted with batch size equal to 1, the reported throughput values are equivalent to TBT.

\begin{table}[t]
  \centering
  \small
  \setlength{\tabcolsep}{5pt}
  \renewcommand{\arraystretch}{1.1}
    \caption{Effieicny Comparision with Throughput}

  \begin{tabular}{lcc}
    \hline
    \textbf{Method} & \textbf{Input Length} & \textbf{Token / Sec)} \\
    \hline
Fully Recompute & 5k & 12.681 \\
SemShareKV      & 5k & 34.065 \\
SnapKV          & 5k & 22.576 \\
H2O             & 5k & 16.273 \\
PyramidKV       & 5k & 3.991  \\
    \hline
  \end{tabular}
  \label{tab:throughput_compare_table}
\end{table}

\begin{figure}[t]
  \includegraphics[width=0.9\columnwidth]{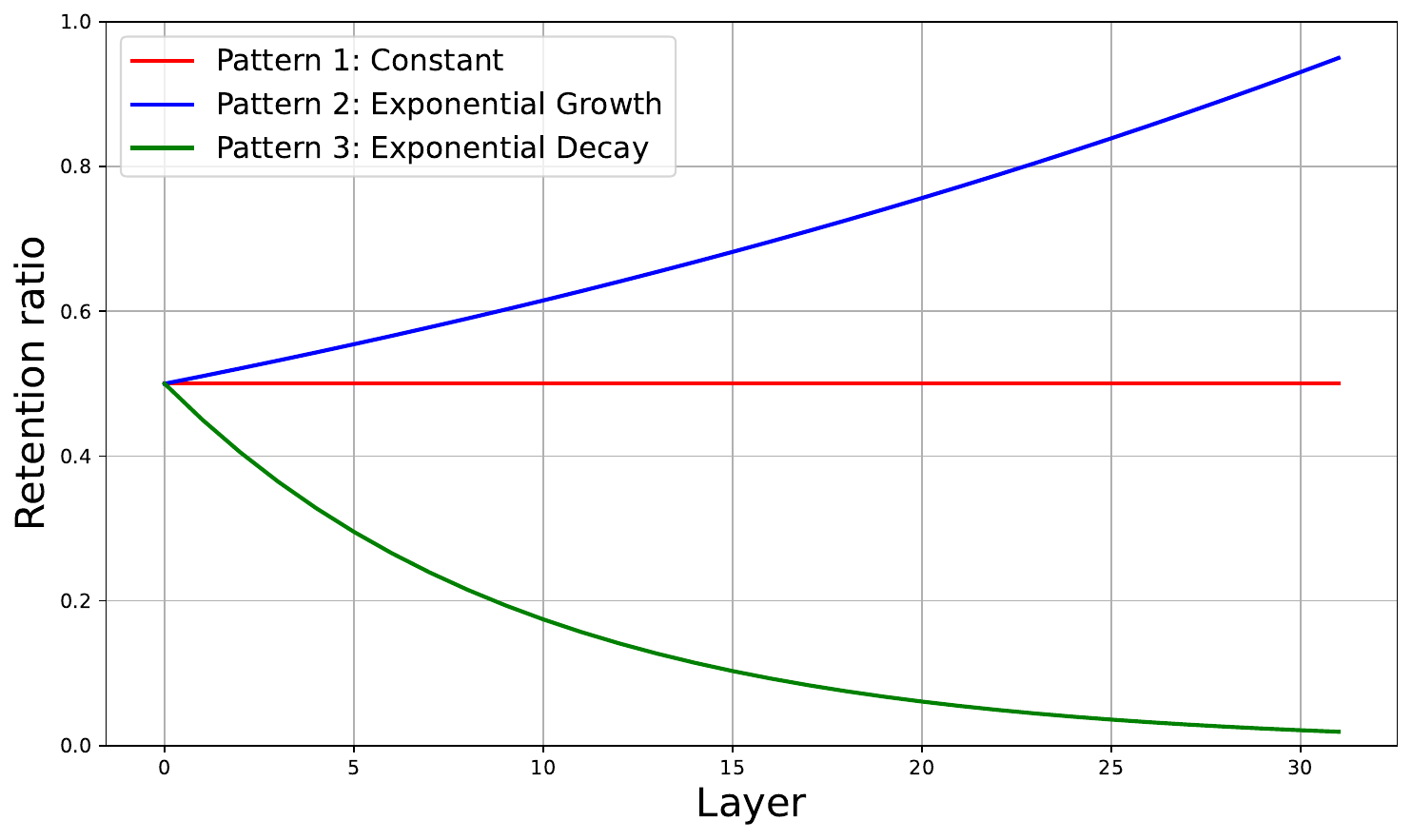}
  \caption{The three retention patterns start from the same retention ratio.}
  \label{fig:insight3}
\end{figure}

\section{Implementation and Hyperparameters}
\label{app:implementation}

SemShareKV is implemented in Python using the \texttt{transformers} library~\citep{wolf-etal-2020-transformers}, with the monkeypatching technique. We use the Locality-Sensitive Hashing from FAISS~\cite{douze2024faiss} library. The code is available at: \url{https://github.com/JasperZhao666/SemShareKV-public.git}. Details of the key functions and their roles are outlined below:

\begin{itemize}
    \item \textbf{mistral\_attn\_forward}: A modified version of \texttt{MistralAttention.forward} from the \texttt{transformers} library, incorporating the SemShareKV mechanism. The hyperparameters used in our experiments are also specified in this function.
    
    \item \textbf{replace\_mistral\_forward}: Applies monkey-patching to substitute the original Mistral model attention forward function in the \texttt{transformers} library with our customized SemShareKV implementation.

    \item \textbf{llama\_attn\_forward}: A modified version of \texttt{LlamaAttention.forward} from the \texttt{transformers} library, incorporating the SemShareKV mechanism. The hyperparameters used in our experiments are also specified in this function.

    \item \textbf{replace\_llama\_forward}: Applies monkey-patching to substitute the original Llama model attention forward function in the \texttt{transformers} library with our customized SemShareKV implementation.
    
    \item \textbf{prepare\_fuzzy\_caches}: Encodes ROPE into E caches and performs fuzzy token matching using locality-sensitive hashing (LSH).
\end{itemize}

In general, SemShareKV is built on the transformer architecture and consists of fewer than 300 new lines of code, which makes it lightweight and easily transferable to other LLMs. 

\section{Artifact Use and Compliance with Intended Purpose}
The datasets used in this study are publicly available and are consistent with their intended use, as specified by the respective sources. In preparing the data, we adhered to ethical guidelines and ensured that the use of these publicly released datasets was for research purposes only.

For the created artifacts, such as the semantically similar samples, we have ensured that the use of these modified datasets remains consistent with the original intended research purpose. The generated data serves the purpose of advancing research in semantic similarity and does not extend beyond the intended scope of the original datasets.
\end{document}